\documentclass[journal]{IEEEtran}
\usepackage{cite}
\usepackage{amsfonts}
\usepackage{subfigure}
\usepackage{multirow}
\usepackage{xcolor}
\usepackage{mathrsfs}
\usepackage{amsmath}
\usepackage{amsthm}
\usepackage{url}
\usepackage{graphicx}
\usepackage{booktabs}
\usepackage{amssymb}
\usepackage{balance}
\usepackage[pagebackref=true,breaklinks=true,letterpaper=true,colorlinks,bookmarks=false]{hyperref}

\ifCLASSINFOpdf
\else
\fi
\begin{document}
%
\title{Occlusion-aware Unsupervised Learning of Depth from 4-D Light Fields}
%
%
%

\author{Jing Jin, 
        and Junhui Hou,~\IEEEmembership{Senior Member, IEEE}
\thanks{This work was supported in part by the Hong Kong Research Grants Council
under grants 9048123 (CityU 21211518) and 9042820 (CityU 11219019),
and in part by the Basic Research General Program of Shenzhen Municipality
under grants JCYJ20190808183003968.}
\thanks{J. Jin and J. Hou are with the Department of Computer Science, City University of Hong Kong, Hong Kong
(e-mail: jingjin25-c@my.cityu.edu.hk;~jh.hou@cityu.edu.hk).}
}

%
%

\markboth{}
{Shell \MakeLowercase{\textit{et al.}}: Bare Demo of IEEEtran.cls for IEEE Journals}
%



\maketitle

\begin{abstract}
Depth estimation is a fundamental issue in 4-D light field processing and analysis.  Although recent supervised learning-based light field depth estimation methods have significantly improved the accuracy and efficiency of traditional optimization-based ones, these methods rely on the training over light field data with ground-truth depth maps which are challenging to obtain or even unavailable for real-world light field data.
Besides, due to the inevitable gap (or domain difference) between real-world and synthetic data, they may suffer from serious performance degradation when generalizing the models trained with synthetic data to real-world data. By contrast, we propose an unsupervised learning-based method, which does not require ground-truth depth as supervision during training. Specifically, based on the basic knowledge of the unique geometry structure of light field data, we present an occlusion-aware strategy to improve the accuracy on occlusion areas, in which  we explore the angular coherence among subsets of the light field views to estimate initial depth maps, and utilize a constrained unsupervised loss to learn their corresponding reliability for final depth  prediction. Additionally, we  adopt a multi-scale network with a weighted smoothness loss to handle the textureless areas. Experimental results on synthetic data show that our method can significantly shrink the performance gap between the previous unsupervised method and supervised ones, and produce depth maps with comparable accuracy to traditional methods with obviously reduced computational cost. Moreover, experiments on real-world datasets show that our method can avoid the domain shift problem presented in supervised methods, demonstrating the great potential of our method. The code will be publicly available at  \url{https://github.com/jingjin25/LFDE-OccUnNet}.
\end{abstract}


%
\IEEEpeerreviewmaketitle

\section{Introduction}



Depth estimation is a crucial topic in the field of computer vision and image processing, which aims at inferring the scene geometry from 2D images.
Compared with monocular or stereo 2-D images,   
light field (LF) images intrinsically record the scene geometry by capturing both the intensity and direction of the light rays permeating the 3-D scenes \cite{levoy1996lf,ng2005lfcamera}, and thus, provide more accurate cues for depth estimation.
In addition, depth estimation is also a common step in 4-D LF processing and analysis, which  paves the way for subsequent LF applications, 
such as 3-D reconstruction \cite{lfdepth2013scene-kim} and virtual/augment reality \cite{lfapp2015vr,lfapp2017vryu}.

Traditional methods for LF depth estimation mainly design different costs to explore 
the depth cues intrinsically provided by the LF data \cite{lfdepth2017taxonomy}, such as detecting the lines in the epipolar plane images (EPIs) and computing their slopes \cite{lfdepth2012globally-wanner,lfdepth2013scene-kim,lfdepth2016spo-zhang,lfdepth2018multiepi-sheng, lfdepth2018trush-schilling},  matching corresponding pixels among different sup-aperture images (SAIs) \cite{lfdepth2015occlusion-wang,lfdepth2016occlusion-wang,lfdepth2017occ-zhu}, and  finding optimal refocused images \cite{lfdepth2013defocus-tao,lfdepth2016robust-williem,lfdepth2017robust-williem}.
However, the procedure of the cost minimization and the followed global optimization always bring a  heavy burden of computation.
Recent learning-based methods \cite{lfdepth2016convolutional-heber,lfdepth2017neural-heber,lfdepth2018epinet-shin,lfdepth2018intrins-anna,lfdepth2020attention-tsai,lfdepth2019framework-shi,lfdepth2021attmlf-chen} significantly improve both the efficiency and accuracy of LF depth estimation by training deep neural networks under the supervision of ground-truth depth maps.
As ground-truth depth maps are usually unavailable for real-world LF images, these methods are always trained using synthetic data that are costly to obtain.
Moreover, due to the inevitable domain shift between the synthetic and real-world data, the models trained with synthetic data always suffer from  performance degradation when generalizing to real-world images. Unsupervised learning-based methods \cite{lfdepth2018unsupervised-peng, lfdepth2020zero-peng, lfdepth2020monocular-zhou} overcome these limitations by training deep neural networks without the need of ground-truth depth maps as supervision. However, their   performance 
is still limited.


Occlusion is a fundamental issue in LF depth estimation and has been being studied these years. Generally, different occlusion-aware LF depth estimation methods are built upon the same fact or similar observations, but differ from each other in the explicit realization and formulation of the fact/observations, resulting in various estimation accuracy and efficiency.
Non-learning-based methods build cost volumes by checking photo-consistency to estimate depth maps from LFs, and then, handle occlusions based on a common fact that the photo-consistency can only hold on non-occluded pixels.
Therefore, previous non-learning-based methods typically design different occlusion-aware cost volumes to improve the accuracy of depth estimation on occlusion areas \cite{lfdepth2015occlusion-wang,lfdepth2017occ-zhu,lfdepth2018trush-schilling,lfdepth2017robust-williem}, which relies in time-consuming cost-volume optimization.
The learning-based method in \cite{lfdepth2020monocular-zhou} also considers the occlusion issue. Following previous unsupervised methods for stereo or optical flow estimation, it uses left-right or forward-backward depth consistency for occlusion detection. However, to check the depth consistency for computing the symmetry loss, the depth maps of the eight SAIs neighboring to the central one need to be predicted simultaneously, which increases computational complexity.

    \begin{figure}[!t]
    \centering
    \includegraphics[width=\linewidth]{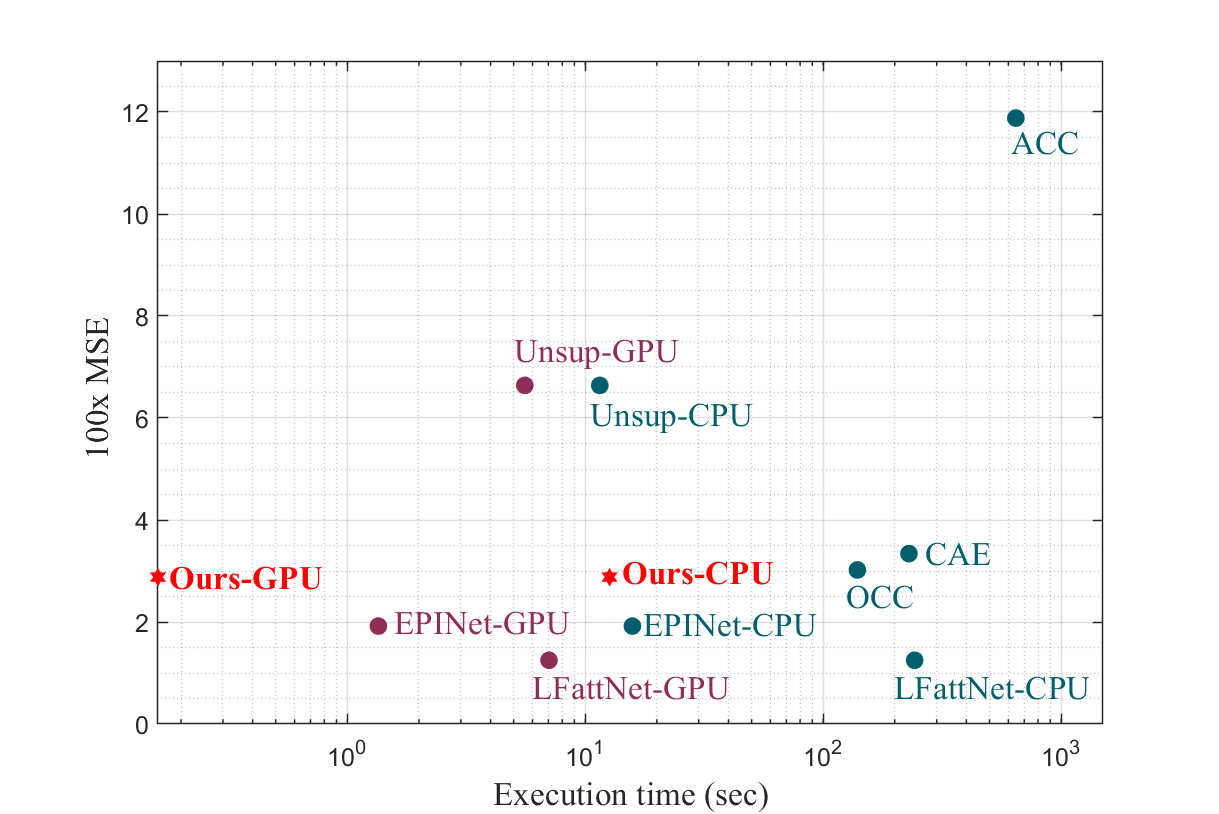}
    \caption{Comparisons of the running time (in second) and the accuracy of the depth estimation of different methods (in MSE, Mean Square Error).  
     For the learning-based methods, including EPINet \cite{lfdepth2018epinet-shin}, LFattNet \cite{lfdepth2020attention-tsai}, Unsup \cite{lfdepth2018unsupervised-peng}, and Ours, we provide the results with and without the GPU acceleration, denoted as GPU and CPU, respectively, while
     for non-learning-based methods, including ACC \cite{lfdepth2015accurate-jeon}, OCC \cite{lfdepth2015occlusion-wang},  and CAE \cite{lfdepth2017robust-williem}, we only evaluated them without the GPU acceleration.
    }
    \label{fig_timemse}
    \end{figure}
    
In this paper, we propose an unsupervised learning-based method for LF depth estimation, which overcomes the limitations of supervised learning-based  and optimization-based methods. That is, the proposed method can not only be trained without using the ground-truth depth as supervision but also infer the depth at a high speed.
In contrast to supervised learning-based methods that can directly regress the depth map under the supervision of the ground-truth depth maps, 
it is challenging for unsupervised learning-based methods to predict the accurate depth map purely from the LF image,  especially on occlusion and textureless areas.
Therefore, we develop the network and loss function based on the knowledge of the unique geometry structure of LF data, 
which is also the general path for designing unsupervised learning methods \cite{wu2020symmetric,guo2020zero}.
Specifically, based on the observation that the angular coherence always holds on a subset of the SAIs in an LF even for pixels with occlusions, we propose a simple yet effective strategy to handle the occlusion issue. That is, instead of processing a 4-D LF as a whole, we partition it into sub-LFs, each of which contains a certain subset of SAIs,  and predict multiple depth maps from sub-LFs. Meanwhile, we design a constrained unsupervised loss to simultaneously predict their reliability maps that are utilized to produce the final prediction.
Additionally, we adopt a multi-scale network together with a global smoothness loss to propagate the depth estimation to textureless areas.

Experimental results on both synthetic and real-world 4-D LF datasets demonstrate the great potential of the proposed method.
Specifically,
our method significantly improves the accuracy of the state-of-the-art unsupervised learning-based one \cite{lfdepth2018unsupervised-peng}, and compared with traditional optimization-based methods, our method achieves comparable even better performance at a much higher speed.
Moreover, our method can avoid the domain shift problem presented in supervised learning-based methods and still produce satisfactory results in real-world LF data.

    \begin{figure*}[!t]
    \centering
    \includegraphics[width=\linewidth]{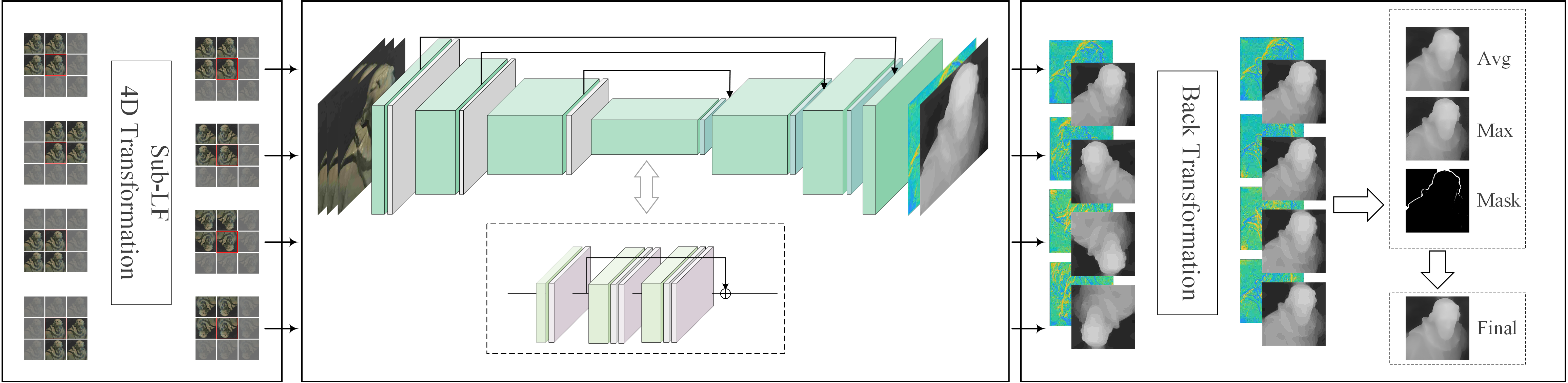}
    \caption{Flowchart of the proposed unsupervised learning-based method for depth estimation from 4-D LFs. The sub-LFs generated from the LF image are first transformed, and then fed into a multi-scale network to produce the initial depth maps and the corresponding reliability maps simultaneously, which are further back transformed to match the central SAI, and fused to produce the final prediction of the depth map. }
    \label{fig_framework}
    \end{figure*}
    
The rest of this paper is organized as follows. Sec. \ref{sec:related_work} comprehensively reviews existing methods for depth estimation from LFs. Sec. \ref{sec:method} describes the proposed method in detail.  Sec. \ref{sec:experiment} presents extensive experimental results to evaluate and further analyse the proposed method. Finally, Sec. \ref{sec:conclusion} concludes this paper.

\section{Related Work}
\label{sec:related_work}
We classify the existing depth estimation methods from LFs into two categories, i.e., non-learning-based and learning-based methods.

\subsection{Non-learning-based Methods}



Traditional non-learning-based methods typically involve two steps, i.e., local depth initialization and global optimization.
These methods explore the geometric property of the LF structure to design different data costs for local depth estimation, mainly including EPI-based and angular patch-based methods.

In the EPI of an LF image, the projections of the same scene point captured by different SAIs construct a straight line, and the slope of the line reflects the depth of the scene point.
Based on this observation, Bolles \textit{et al.} \cite{bolles1987epipolar} first proposed to detect the straight lines in EPIs and compute their slopes for depth estimation.
Wanner \textit{et al.}  \cite{lfdepth2012globally-wanner} used the structure tensor to compute the direction of the local level lines in EPIs and used the coherence of the structure tensor to measure the reliability.
Kim \textit{et al.} \cite{lfdepth2013scene-kim}  first estimated the slope of line edges on EPIs, and then propagated the information throughout the EPI. They also used a fine-to-coarse strategy to successively process the EPI for the highest resolution to the coarser resolution.
Zhang \textit{et al.} \cite{lfdepth2016spo-zhang} proposed a spinning parallelogram operator to obtain local depth estimation, which defines a parallelogram on the EPI and figures out the straight line by finding the maximum distance between the distributions of pixel values on either sider of the lines.

Angular patch-based methods extend conventional stereo matching to narrow-baseline multi-view geometry, and construct different costs to measure the photo-consistency among different SAIs of the LF.
Jeon \textit{et al.} \cite{lfdepth2015accurate-jeon}  applied the phase shift theorem to estimate the sub-pixel displacement between SAIs of the LF, and built gradient-based and feature-based matching costs to find correspondences.
Some methods also introduced the refocus image as the depth cue \cite{lfdepth2013defocus-tao,lfdepth2015shading-tao, lfdepth2015focal-lin}.
The assumption of photo-consistency does not hold in the presence of occlusions, and thus, different methods have been proposed to address this issue. 
Wang \textit{et al.} \cite{lfdepth2015occlusion-wang,lfdepth2016occlusion-wang} developed the LF occlusion theory, i.e.,  the line separating the view regions in the angular patch has the same orientation as the occlusion edge in the spatial domain, and then computed depth on half of the viewpoints that keep the photo-consistency.
Zhu \textit{et al.} \cite{lfdepth2017occ-zhu} further extended this occlusion theory to the case of multi-occluder occlusion.
Chen \textit{et al.} \cite{lfdepth2018pobr-chen} defined the partially occluded border regions which suffer from occlusion induced depth uncertainty, and employed superpixel-based regularization to resolve such uncertainty.
Williem \textit{et al.} \cite{lfdepth2016robust-williem,lfdepth2017robust-williem} proposed an angular entropy cost and an adaptive defocus response to improve the robustness on occlusions and noisy.
Instead of utilizing partial angular patches for depth estimation, Zhang \textit{et al.} \cite{lfdepth2020graph-zhang} leveraged the undirected graph to jointly consider occluded and unoccluded SAIs in the angular patch to exploit the structural information of the LF.
Note that in parallel to our work, one contemporaneous
work \cite{lfdepth2021votecost-han} has emerged recently, 
where the occlusion issue is addressed by proposing a novel vote cost to count the number of refocused pixels whose deviations from the central-view pixel are less than a small threshold.

\subsection{Learning-based Methods}

Through data-driven training with the ground-truth depth as supervision,
recent learning-based methods significantly improve the depth estimation accuracy while greatly saving the computational time.

Heber \textit{et al.} \cite{lfdepth2016convolutional-heber} applied a 5-layers network to  two image patches from the vertical and horizontal EPIs to predict the depth value for individual pixels. They further improved the methods by using a u-shaped network with 3-D convolutional layers to predict the depth map for the full central view through one pass forward \cite{lfdepth2017neural-heber}.
Shin \textit{et al.} \cite{lfdepth2018epinet-shin} proposed a multi-stream network to extract features from SAI stacks with 4 directions, and then concatenate the features to regress the depth map.
Shi \textit{et al.} \cite{lfdepth2019framework-shi} proposed a learning-based framework for depth estimation, which obtains the initial depth map for stereo pairs using fine-tuned model of FlowNet2.0 optical flow estimation \cite{flow2015flownet-dosovitskiy,flow2017flowne2-tilg}, and then fuses the initial depth estimations based on the warping errors to handle occlusions.
Most recently,
Tsai \textit{et al.} \cite{lfdepth2020attention-tsai} introduced an SAI selection module to generate attention maps that indicate the importance of each SAI,
and Chen \textit{et al.} \cite{lfdepth2021attmlf-chen} also used the attention mechanism to fuse features from different SAIs.
To handle the occlusion issue, 
Guo \textit{et al.} \cite{lfdepth2020occlusion-guo}   computed the occlusion regions based on the groun-truth depth maps, and utilized a network for occlusion detection in a supervised manner. Then they estimate the depth map in occlusion and non-occlusion regions separately.

All the above-mentioned learning-based methods have to be trained with ground-truth depth. However, the ground-truth depth maps are challenging to obtain or even unavailable for real-world LF data. Besides, they may suffer from performance degradation when generalizing the models trained with synthetic data to real-world data due to the domain difference between synthetic and real-world data.
Recently, Peng \textit{et al.} \cite{lfdepth2018unsupervised-peng,lfdepth2020zero-peng} proposed an unsupervised learning method for depth estimation from LFs, which can be trained without the ground-truth depth maps. 
However, as the occlusion and textureless issues are not expelicitly addressed, the accuracy of the estimated depth maps is still limited.
Apart from the similar photometric consistency-based losses, Zhou \textit{et al.} \cite{lfdepth2020monocular-zhou} proposed a symmetry loss to handle occlusion areas.
However, to check the depth consistency for computing the symmetry loss, the depth maps for the eight SAIs neighboring to the central one need to be predicted simultaneously, which increases the computational complexity.



\section{Proposed Method}
\label{sec:method}
\subsection{Overview}
Let $L(x,y,u,v)\in\mathbb{R}^{H\times W\times M\times N}$ denote a 4-D LF image with the spatial resolution of $H\times W$ and the angular resolution of $M\times N$.
The LF image can be regarded as a set of 2-D views observed from the viewpoints distributed on a 2-D plane, and thus, it can also be denoted by $\mathcal{L}=\{I_{\mathbf{u}} = L(:,:,u,v)\in\mathbb{R}^{H\times W} | \mathbf{u}\subset \mathcal{U}\}$, where $I_{\mathbf{u}}$ denotes the SAI at the angular position $\mathbf{u}=(u,v)$, and $\mathcal{U}$ is the set of the 2-D angular coordinates, i.e.,
$\mathcal{U}=\{\mathbf{u} | \mathbf{u} = (u,v), 1\leq u\leq M, 1\leq v\leq N \}$.
Let $I_{\mathbf{u}_0}$ denote the central view of $\mathcal{L}$, 
and our goal is to estimate its depth map, denoted as $D$. Note that as the depth is inversely proportional to the disparity, we do not make a distinction between them in this paper.


Considering the limitations of supervised learning-based methods and traditional optimization-based methods, we aim at estimating $D$ using an unsupervised learning-based framework.
With no ground-truth labels as the supervision during training, the general guidance for designing an unsupervised learning method is to take advantage of the prior knowledge of the data \cite{wu2020symmetric}. 
In our method, we exploit the intrinsic geometry structure of LF data to drive the training of the network.
To be more specific, based on the observation that the photo consistency can always hold within a subset of the LF views, we proposed an occlusion-aware strategy to address the occlusion issue, i.e., we predict the initial depth maps from sub-LFs and estimate their occlusion-aware reliability using a constrained unsupervised loss.
The final prediction of the depth map can be produced by fusing the initial depth maps based on their reliability maps.
Additionally, we build the network in a multi-scale manner and adopt an edge-weighted smoothness loss to propagate the depth estimation to textureless areas.

Fig. \ref{fig_framework} illustrates the framework of our proposed method.
In what follows, we will first introduce the general idea for unsupervised depth estimation and then describe our method in detail.

\subsection{Unsupervised LF Depth Estimation}
The main cue for the depth estimation from $\mathcal{L}$ is the angular coherence among the views, i.e., the projections of the same scene point at different views have the same intensity under the assumption of Lambertian and non-occlusion. This relation can be represented as:
\begin{equation}
      I_{\mathbf{u}_0}(\mathbf{x}) = I_\mathbf{u}(\mathbf{x}+D(\mathbf{x})(\mathbf{u}-\mathbf{u}_0)),
\label{eq_lfstructure}
\end{equation}
where $\mathbf{x}$ is the spatial coordinate.
Based on Eq. (\ref{eq_lfstructure}), 
we can build 
a feature-extraction network
to explore the angular coherence among the LF for depth estimation. 
Moreover, we can use an unsupervised loss to train the network by minimizing the photometric reconstruction distance,
i.e.,
\begin{equation}
      \ell_{rec} (\mathcal{L},\widetilde{D}) =  \sum_{\mathbf{u}\subset\mathcal{U}}  \sum_{\mathbf{x}}  \left| \widehat{I} _{\mathbf{u}\to\mathbf{u}_0} \left(\mathbf{x};\widetilde{D}\right) - I_{\mathbf{u}_0} (\mathbf{x}) \right|,
\label{eq_unsuploss}
\end{equation}
where $\widehat{I} _{\mathbf{u}\to\mathbf{u}_0} \left(\widetilde{D}\right) $ denotes the image warped from $I_\mathbf{u}$ to $I_{\mathbf{u}_0}$ based on  the predicted depth map $\widetilde{D}$.

However,  the relationship described in Eq. (\ref{eq_lfstructure}) no longer holds when occlusions occur, and thus, the model trained with the loss function in Eq. (\ref{eq_unsuploss}) will lose accuracy on occlusion areas.
To this end, we propose the following occlusion-aware strategy for unsupervised learning of the LF depth estimation.


    \begin{figure}
    \centering
    \subfigure[The imaging model when the occlusion occurs]{
    \begin{minipage}[b]{\linewidth}
    \includegraphics[width=\linewidth]{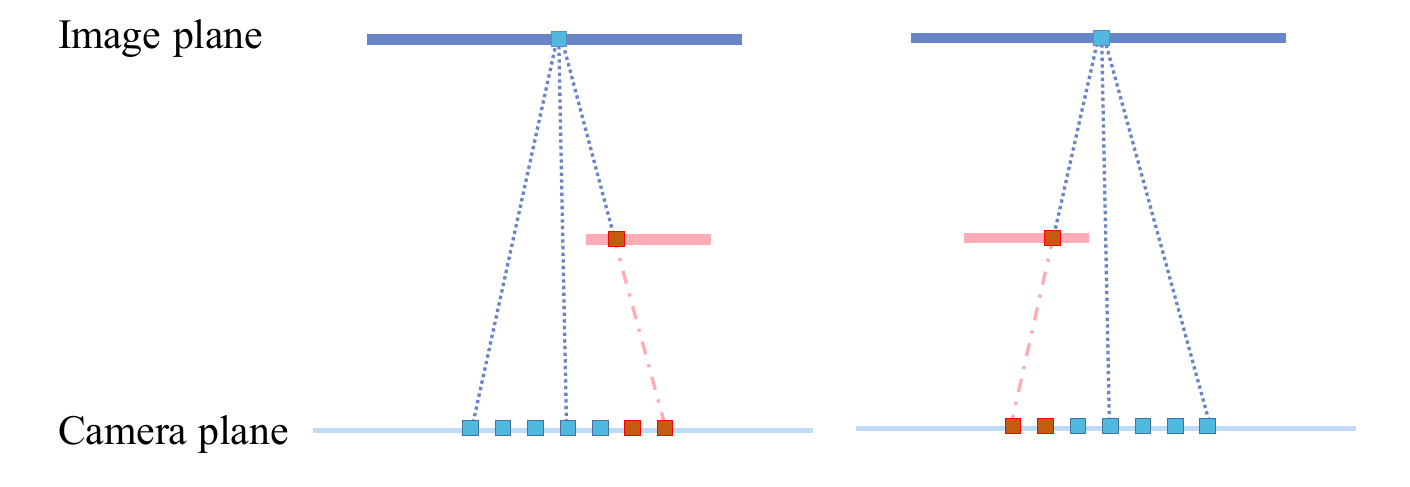} 
    \end{minipage}
    }
    \subfigure[Appearances of the spatial and angular patches of pixels with occlusions]{
    \begin{minipage}[b]{\linewidth}
    \includegraphics[width=\linewidth]{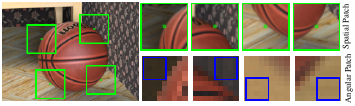}
    \end{minipage}
    }
    \caption{Illustration of the occlusion model. In (a), the blue dots are the scene points on the image plane and their projections on the camera plane, and red dots are the occluders and their projections. In (b), the green dots are pixels with occlusions from the central view, and their surrounding areas, namely spatial patches, are framed in green and zoomed in. The refocused angular patch of each green dot with respect to its ground-truth depth value is presented under its spatial patch, where the blue frame indicates the part of the angular patch that obeys the photo consistency. 
    }
    \label{fig_ang_patch}
    \end{figure}

\subsection{Occlusion-aware Model}
We can use the refocused angular patch to represent the  unique structure of LF data described in Eq. (\ref{eq_lfstructure}) in a more straightforward way. 
The refocused angular patch for the pixel $\mathbf{x}$ of $I_{\mathbf{u}_0}$ with respect to the depth value $d$ is denoted by $A_{\mathbf{x},d} (u,v) = L(x+d(u - u_0), y+d(v-v_0),u,v)$.
$A_{\mathbf{x},d}$ collects the corresponding pixels of $I_{\mathbf{u}_0}(\mathbf{x})$ from each SAIs of the LF refocused at depth $d$. 
Based on Eq. (\ref{eq_lfstructure}), pixels in $A_{\mathbf{x},d}$ will have the same intensity when $d$ is the correct depth value of $I_{\mathbf{u}_0}(\mathbf{x})$, i.e., $d=D(\mathbf{x})$.
However, when the occlusion occurs at $I_{\mathbf{u}_0}(\mathbf{x})$, the photo consistency  will not hold in $A_{\mathbf{x},D(\mathbf{x})}$.

Fig. \ref{fig_ang_patch} (a) illustrates the imaging model when the occlusion occurs.
Suppose only one occluder exists, the occlusion will only occur at one side of the central viewpoint along a 1-D angular dimension.
By extending this observation to the 2-D angular plane of an LF image, we can deduce that 
one of the 4 sides of the central viewpoint can avoid the occlusion problem.
As shown in  Fig. \ref{fig_ang_patch} (b), we demonstrate 4 different occlusion scenarios (as shown in the spatial patches framed in green), and collect the refocused angular patches of the occluded pixels  using their ground-truth depth values.
We can observe that  although the whole angular patches disobey the color consistency, if we divide $A_{\mathbf{x}, D(\mathbf{x})}$ into 4 parts, then at least one part can still keep the color consistency (as shown in the blue frames).
Moreover, the positions of the color-consistent part differ with the spatial and geometric content in the spatial patches.

Based on such an unique structure of LF data, we propose a simple yet effective strategy to handle occlusions for the LF depth estimation. Specifically, 
we estimate depth maps from sub-LFs instead of the full LF, and predict their reliability maps simultaneously.
Then, we obtain the final prediction of the depth map by fusing those from sub-LFs under the guidance of their reliability maps.

\subsubsection{Sub-LF generation}
We first divide each of the angular dimension with $\mathbf{u}_0 = (M_0, N_0)$ as the center to get 4 sub-sets of $\mathcal{U}$, denoted as $\mathcal{U} = \left\{\mathcal{U}_1, \mathcal{U}_2, \mathcal{U}_3, \mathcal{U}_4 \right\}$. 
Note that $\mathbf{u}_0$ is the angular position of the depth map to be estimated.
Specifically, $\mathcal{U}_1 = \{\mathbf{u}=(u,v)| 1\leq u\leq M_0, 1\leq v\leq N_0\}$,  $\mathcal{U}_2 = \{\mathbf{u}=(u,v)| 1\leq u\leq M_0, N_0\leq v\leq N\}$,  $\mathcal{U}_3 = \{\mathbf{u}=(u,v)| M_0\leq u\leq M, 1\leq v\leq N_0\}$,   and $\mathcal{U}_4 = \{\mathbf{u}=(u,v)| M_0\leq u\leq M, N_0\leq v\leq N\}$.
Accordingly, we divide the LF image into 4 sub-LFs, i.e., $\mathcal{L}=\{\mathcal{L}_1, \mathcal{L}_2, \mathcal{L}_3, \mathcal{L}_4\}$, where $\mathcal{L}_i = \{I_\mathbf{u} | \mathbf{u}\subset \mathcal{U}_i\}$.

\subsubsection{Uncertainty-aware depth maps}
We apply a network to predict the depth map from these sub-LFs, producing $\widetilde{D}_1, \widetilde{D}_2, \widetilde{D}_3$, and $\widetilde{D}_4$.
Each sub-LF contains structured views to provide sufficient cues for depth estimation, but will fail when the occlusion occurs. 
Fortunately, most pixels of $I_{\mathbf{u}_0}$ have at least one sub-LF that maintains the angular coherence.
To indicate which sub-LF is reliable for depth estimation, we expect  the network to learn a reliability map for each prediction of the sub-LF simultaneously.
To enable the training of such a network, we propose a constrained photometric reconstruction loss defined as:
\begin{equation}
\begin{aligned}
      &\ell_{c-rec} (\mathcal{L},\widetilde{D}, \widetilde{W}) =\\
      &\sum_{i=1}^4 \sum_{\mathbf{u}\subset\mathcal{U}_i} \sum_{\mathbf{x}} \widetilde{W}_i (\mathbf{x})  \left| \widehat{I}_{\mathbf{u}\to\mathbf{u}_0} \left(\mathbf{x}; \widetilde{D}_i \right) - I_{\mathbf{u}_0} (\mathbf{x}) \right|,      
\end{aligned}
\label{eq_cons_recloss}
\end{equation}
where 
$\widetilde{W}_i$ is the reliability map corresponding to $\widetilde{D}_i$.  We also apply the softmax operation across different uncertainty maps to ensure $\sum_{i=1}^4 \widetilde{W}_i (\mathbf{x})= 1$.

\begin{figure}[!t]
    \centering
    \includegraphics[width=\linewidth]{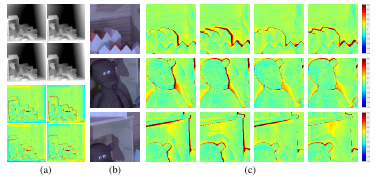}
    \caption{Visual illustration of the uncertainty-aware depth maps estimated from the sub-LFs. (a) The depth maps estimated from sub-LFs  and their error maps with respect to the ground-truth depth; (b) Zoom-in spatial patches; (c) Zoom-in error maps of the  depth maps estimated from sub-LFs
    (the results from $\mathcal{L}_1$, $\mathcal{L}_2$, $\mathcal{L}_3$, and $\mathcal{L}_4$ are presented from left to right.)
    }
    \label{fig_sub_depth}
    \end{figure}

By minimizing  $\ell_{c-rec}$ instead of $\ell_{rec}$, the network can relax the optimization on occlusion areas of each sub-LF, 
because the loss can be decreased by reducing $\widetilde{W}_i (\mathbf{x}) $ when $ \widehat{I}_{\mathbf{u}\to\mathbf{u}_0} \left(\mathbf{x}; \widetilde{D}_i \right) $ has difficulty to get close to $ I_{\mathbf{u}_0} (\mathbf{x})$.
Fig. \ref{fig_sub_depth} visually illustrates its effect by showing the error maps of the initial depth maps estimated from sub-LFs, where it can be observed that in each spatial patch with occlusions, some of the depth maps show high accuracy while others contain obvious errors. Moreover, the subset of the accurate depth maps changes with the spatial content and geometric relations of patches.

    \begin{figure}[!t]
    \centering
    \includegraphics[width=1.0\linewidth]{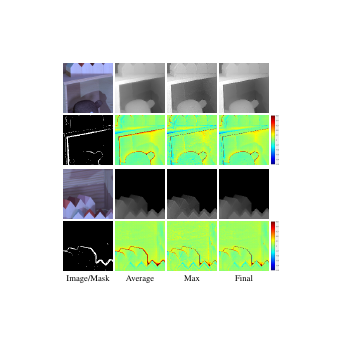}
    \caption{Visual illustration of the occlusion-aware fusion. The image patch of the central view, the occlusion mask, the depth maps from average and max operations, and the final prediction of the depth maps are presented. The corresponding error map is also provided under each depth map.}
    \label{fig_occ_fusion}
    \end{figure}
    
\subsubsection{Occlusion-aware fusion}
We generate the final prediction of the depth map by fusing the 4 initial depth maps under the guidance of 
their reliability maps.
For pixels in occlusion areas, only the sub-LFs without any  occluded viewpoints can hold the angular coherence and provide accurate depth estimation. Therefore, we take the depth value with the highest reliability as the final prediction, i.e.,
\begin{equation}
      \widetilde{D}_{max} (\mathbf{x}) = \widetilde{D}_j (\mathbf{x}), \quad j = \mathop{\arg\max}_i \widetilde{W}_i(\mathbf{x}).
\label{eq_depth_max}
\end{equation}
Considering that the reliability map in Eq. \eqref{eq_cons_recloss} may increase the uncertainty of the non-occlusion pixels of the initial depth maps from individual sub-LFs,
we take the average depth values weighted by the reliability to produce a smooth estimation,
i.e.,
\begin{equation}
      \widetilde{D}_{avg} (\mathbf{x}) =\sum_i \widetilde{W}_i(\mathbf{x})  \widetilde{D}_j (\mathbf{x}).
\label{eq_depth_avg}
\end{equation}
The detection of the occlusion areas is based on the variance of the initial depth maps.
As all the 4 sub-LFs on non-occlusion areas obey the intensity consistency when refocusing on the correct depth, they can produce relatively accurate predictions that  are close to each other. 
While on occlusion areas, some of the sub-LFs are influenced  by the intensity inconsistency, and others are not, and thus, the predictions show a relatively large divergence.
We compute the standard deviation over the initial prediction for each pixel, and make a hard threshold to generate the occlusion mask, i.e., 
\begin{equation}
      \widetilde{O}  (\mathbf{x}) = \mathsf{T}_\lambda \left(\mathsf{STD}\left(\widetilde{D}_1,\widetilde{D}_2,\widetilde{D}_3,\widetilde{D}_4\right)\right),
\label{eq_occ_mask}
\end{equation}
where 
$\mathsf{T}_\lambda (x) =\left\{
\begin{aligned}
1,  \quad & x \geq \lambda \\
0,  \quad & x < \lambda
\end{aligned}
\right.
$ is the binarization operation with the threshold $\lambda$, 
$\mathsf{STD}(\cdot)$ copmutes the standard deviation, and $\widetilde{O}$ is the estimated occlusion mask with 1 and 0 respectively indicating occlusion and non-occlusion. We empirically set the value of $\lambda$ to 0.3.
Finally, we can produce the final prediction of the depth map at $I_{\mathbf{u}_0}$ by fusing the results on occlusion and non-occlusion areas, i.e.,
\begin{equation}
      \widetilde{D}  (\mathbf{x}) =  \widetilde{O} (\mathbf{x}) \widetilde{D}_{max} (\mathbf{x}) + (1-\widetilde{O} (\mathbf{x}))\widetilde{D}_{avg}.
\label{eq_depth}
\end{equation}

As an example, Fig. \ref{fig_occ_fusion} visualizes  $\widetilde{D}_{max}$, $\widetilde{D}_{avg}$, $\widetilde{O}$, and $\widetilde{D}$ to illustrate the effectiveness of the occlusion-aware fusion, where we can observe that $\widetilde{D}_{max}$ performs well on occlusion boundaries, while $\widetilde{D}_{avg}$ can produce smooth results on non-occlusion areas. Based on the occlusion mask, $\widetilde{D}$ can leverage the advantages of  $\widetilde{D}_{max}$ and $\widetilde{D}_{avg}$.

    \begin{table*}[t]
    \centering
    \caption{ Quantitative comparisons (MSE $\times 100$) of the depth estimation results from different methods on synthetic LF data. The smaller, the better.
    }
    \label{tab:quan_mse}
    \resizebox{\textwidth}{!}{
    \begin{tabular}{l  l | c c c c | c c c c c c}
    \toprule[1pt]
    \multicolumn{2}{c|}{\multirow{2}{*}{Methods}}
     & \multicolumn{4}{c|}{HCI} & \multicolumn{6}{c}{HCIold} \\
    ~ & ~ & Boxes & Cotton & Dino & Sideboard & Buddha & Horses & Medieval & Monasroom & Papillon & StillLife \\
    \midrule[0.5pt]
    \multirow{3}{*}{Non-Learning} & ACC \cite{lfdepth2015accurate-jeon} & 24.91 &	8.70&	1.25&	12.64&	1.21&	1.74&	1.05&	11.02&	4.88&	13.07 \\
    ~ & OCC \cite{lfdepth2015occlusion-wang} &  8.14&	1.04&	0.59&	2.31&	0.76&	0.70&	0.85&	0.55&	0.75&	2.49 \\
    ~ & CAE \cite{lfdepth2017robust-williem} & 10.18&	1.01&	0.62&	1.55&	0.77&	1.20&	1.14&	0.68&	0.85&	1.45 \\
    \midrule[0.5pt]
    \multirow{2}{*}{Supervised}& EPINet \cite{lfdepth2018epinet-shin} & 6.35&	0.26&	0.18&	0.88&	0.37&	6.85&	2.23&	1.36&	6.15&	2.55 \\
    ~ & LFattNet \cite{lfdepth2020attention-tsai} & 4.17&	0.19&	0.10&	0.54&	0.36&	5.79&	1.45&	0.75&	5.15&	15.36\\
    \midrule[0.5pt]
    \multirow{2}{*}{Unsupervised} & Unsup \cite{lfdepth2018unsupervised-peng} & 12.74 & 	7.31& 	1.88 & 	4.62 & 	1.11 & 	1.65 & 	1.27 & 	1.92 & 	4.58 & 	33.90 \\
    ~ & Ours & 7.45	& 0.80 & 	0.63& 	1.79& 	0.34& 	1.52& 	0.70 & 	0.57& 	1.11& 	1.57\\
    \bottomrule[1pt]
    \end{tabular}
    }
    \end{table*}

    \begin{table*}[t]
    \centering
    \caption{ Quantitative comparisons (BPR ($>0.07$)) of the depth estimation results from different methods on synthetic LF data. The smaller, the better.
    }
    \label{tab:quan_bpr}
    \resizebox{\textwidth}{!}{
    \begin{tabular}{l  l | c c c c | c c c c c c}
    \toprule[1pt]
    \multicolumn{2}{c|}{\multirow{2}{*}{Methods}}
     & \multicolumn{4}{c|}{HCI} & \multicolumn{6}{c}{HCIold} \\
    ~ & ~ & Boxes & Cotton & Dino & Sideboard & Buddha & Horses & Medieval & Monasroom & Papillon & StillLife \\
    \midrule[0.5pt]
    \multirow{3}{*}{Non-Learning} & ACC \cite{lfdepth2015accurate-jeon} &25.39&	7.36&	17.57&	24.40&	9.33&	14.44&	7.92&	12.07&	18.14&	22.32
 \\
    ~ & OCC \cite{lfdepth2015occlusion-wang} &  65.68&	4.57&	9.83&	22.92&	6.74&	14.36&	12.74&	10.54&	22.73&	13.06
 \\
    ~ & CAE \cite{lfdepth2017robust-williem} & 29.48&	7.84&	18.17&	21.47&	5.69&	14.82&	23.80&	9.56&	18.59&	20.41
\\
    \midrule[0.5pt]
    \multirow{2}{*}{Supervised}& EPINet \cite{lfdepth2018epinet-shin} & 13.10&	0.47&	1.41&	5.19&	1.62&	16.59&	18.83&	10.56&	36.16&	11.87
 \\
    ~ & LFattNet \cite{lfdepth2020attention-tsai} & 11.51&	0.26&	0.88&	2.97&	2.22&	16.75&	18.64&	9.44&	36.03&	13.02
\\
    \midrule[0.5pt]
    \multirow{2}{*}{Unsupervised} & Unsup \cite{lfdepth2018unsupervised-peng} & 43.82&	28.02&	22.08&	28.06&	9.39&	19.64&	18.41&	14.63&	28.30&	44.97
 \\
    ~ & Ours & 26.24&	8.46&	8.25&	14.20 &	4.11&	26.95&	16.48&	10.57&	36.36&	17.14 \\
    \bottomrule[1pt]
    \end{tabular}
    }
    \end{table*}

\subsection{Implementation Details}

\subsubsection{Sub-LF 4-D transformation}
To reduce the model size, we apply a shared network to the 4 sub-LFs.
As the target view, i.e., the central view of the full LF, is located at different position on the angular plane of each sub-LF, 
we apply a 4-D transformation, which include an angular flip followed by a spatial flip, to the sub-LFs to ensure that they can produce the depth map for the same target.
Specifically, we first take $\mathcal{L}_1$ as the reference and apply an angular flip on the rest of the sub-LFs to adjust the angular position of $I_{\mathbf{u}_0}$.
As the angular flip will destroy the original LF structure \cite{lfsr2020hybrid-jin},  we consequently apply a spatial flip on the corresponding dimensions to recover the LF structure.
Explicitly, the 4-D transformations is written as
\begin{equation}
\begin{aligned}
      \widehat{\mathcal{L}}_1(x,y,u,v) &= \mathcal{L}_1(x,y,u,v),\\
      \widehat{\mathcal{L}}_2(x,y,u,v) &= \mathcal{L}_2(x,W-y,u,N-v),\\
      \widehat{\mathcal{L}}_3(x,y,u,v) &= \mathcal{L}_3(H-x,y,M-u,v),\\
      \widehat{\mathcal{L}}_4(x,y,u,v) &= \mathcal{L}_4(H-x,W-y,M-u,N-v),\\
\end{aligned}
\label{eq_back}
\end{equation}
where $\widehat{\mathcal{L}_i}$ is the transformed sub-LF to be fed into the network for depth estimation.
Additionally, the predicted depth maps will correspond to the flipped $I_{\mathbf{u}_0}$, and thus, they are spatially flipped back to align the same target.
The left and right parts of Fig. \ref{fig_framework} illustrate 
the 4-D transformation applied on the sub-LFs and the back transformation applied on the predicted depth maps, respectively. 

    \begin{table*}[t]
    \centering
    \caption{ Quantitative comparisons (MSE $\times100$) of the depth estimation results from different methods on the HCI 4D benchmark. 
    }
    \label{tab:benchmark_mse}
    \resizebox{\textwidth}{!}{
    \begin{tabular}{l  l | c c c c | c c c c}
    \toprule[1pt]
    \multicolumn{2}{c|}{\multirow{2}{*}{Methods}}
     & \multicolumn{4}{c|}{Test} & \multicolumn{4}{c}{Stratified} \\
    ~ & ~ & Bedroom & Bicycle & Herbs & Origami & Backgammon & Dots & Pyramids & Stripes \\
    \midrule[0.5pt]
    \multirow{3}{*}{Non-Learning} & ACC \cite{lfdepth2015accurate-jeon} & 0.467 & 11.729 & 21.335 & 6.757 & 13.007 & 5.676 & 0.273 & 17.454\\
    ~ & OCC \cite{lfdepth2015occlusion-wang} &  0.633 & 7.669 & 22.202 & 2.300 & 21.587 & 3.301 & 0.098 & 8.131 \\
    ~ & CAE \cite{lfdepth2017robust-williem} & 0.234 & 5.135 & 11.665 & 1.778 & 6.074 & 5.082 & 0.048 & 3.556  \\
    \midrule[0.5pt]
    \multirow{3}{*}{Supervised}& EPINet \cite{lfdepth2018epinet-shin}& 0.213 & 4.682 & 9.700 & 1.466 & 3.629 & 1.635 & 0.008 & 0.950 \\
    ~ & LFattNet \cite{lfdepth2020attention-tsai} & 0.366 & 3.350 & 6.605 & 1.733 & 3.648 & 1.425 & 0.004 & 0.892\\
    ~ & AttMLFNet \cite{lfdepth2021attmlf-chen} & 0.129 & 3.082 & 6.374 & 0.991 & 3.863 & 1.035 & 0.003 & 0.814\\
    \midrule[0.5pt]
    \multirow{3}{*}{Unsupervised} & Unsup \cite{lfdepth2018unsupervised-peng} & 0.924 & 11.737 & 145.551 & 8.817 & 34.709 & 72.998 & 0.035 & 11.759   \\
    ~ & Mono \cite{lfdepth2020monocular-zhou} & 0.415 & 9.232 & 26.816 & 3.679 & 11.833 & 2.536 & 0.027 & 2.677\\
    ~ &  Ours & 0.385 & 6.232 & 13.941 & 1.921 & 6.684 & 6.565 & 0.213 & 5.200\\
    \bottomrule[1pt]
    \end{tabular}
    }
    \end{table*}

    \begin{table*}[t]
    \centering
    \caption{ Quantitative comparisons (BPR ($>0.07$)) of the depth estimation results from different methods on the HCI 4D benchmark.  
    }
    \label{tab:benchmark_bpr}
    \resizebox{\textwidth}{!}{
    \begin{tabular}{l  l | c c c c | c c c c}
    \toprule[1pt]
    \multicolumn{2}{c|}{\multirow{2}{*}{Methods}}
     & \multicolumn{4}{c|}{Test} & \multicolumn{4}{c}{Stratified} \\
    ~ & ~ & Bedroom & Bicycle & Herbs & Origami & Backgammon & Dots & Pyramids & Stripes \\
    \midrule[0.5pt]
    \multirow{3}{*}{Non-Learning} & ACC \cite{lfdepth2015accurate-jeon} & 13.855 & 19.791 & 18.108 & 14.183 & 5.516 & 2.900 & 12.354 & 35.741 \\
    ~ & OCC \cite{lfdepth2015occlusion-wang} & 17.565 & 21.562 & 36.830 & 22.431 & 19.006 & 5.822 & 3.172 & 18.408\\
    ~ & CAE \cite{lfdepth2017robust-williem} & 5.788 & 11.223 & 9.550 & 10.027 & 3.924 & 12.401 & 1.681 & 7.872\\
    \midrule[0.5pt]
    \multirow{3}{*}{Supervised}& EPINet \cite{lfdepth2018epinet-shin}& 2.403 & 9.896 & 12.100 & 5.918 & 3.580 & 3.183 & 0.192 & 2.462 \\
    ~ & LFattNet \cite{lfdepth2020attention-tsai} & 2.792 & 9.511 & 5.219 & 4.824 & 3.126 & 1.432 & 0.195 & 2.933 \\
    ~ & AttMLFNet \cite{lfdepth2021attmlf-chen} & 2.074 & 8.837 & 5.426 & 4.406 & 3.228 & 1.606 & 0.174 & 2.932\\
    \midrule[0.5pt]
    \multirow{2}{*}{Unsupervised} & Unsup \cite{lfdepth2018unsupervised-peng} & 21.604 & 30.239 & 63.940 & 53.408 & 36.419 & 56.102 & 0.809 & 62.864 \\
     ~ & Mono \cite{lfdepth2020monocular-zhou} & 7.413 & 20.098 & 13.443 & 10.949 & 12.311 & 3.651 & 0.262 & 11.136 \\
    ~ &  Ours & 12.687 & 21.650 & 16.959 & 19.821 & 14.371 & 45.340 & 7.348 & 41.359\\
    \bottomrule[1pt]
    \end{tabular}
    }
    \end{table*}

    \begin{table*}[t]
    \centering
    \caption{Comparison of the running time (in seconds) of different methods for inferring the depth map from a $512\times 512\times 7\times 7$ LF image. The inference time with and without GPU acceleration are provided. }
    \label{tab:time}
    \resizebox{0.7\textwidth}{!}{
    \begin{tabular}{c| c c c |c c| c c }
    \toprule[1pt]
    ~ & \multicolumn{3}{c|}{Non-Learning} & \multicolumn{2}{c|}{Supervised} & \multicolumn{2}{c}{Unsupervised}\\ 
    \midrule[0.5pt]
    ~ & ACC \cite{lfdepth2015accurate-jeon} & OCC \cite{lfdepth2015occlusion-wang} & CAE \cite{lfdepth2017robust-williem} & EPINet \cite{lfdepth2018epinet-shin} & LFattNet \cite{lfdepth2020attention-tsai} & Unsup \cite{lfdepth2018unsupervised-peng} &  Ours \\
    \midrule[0.5pt]
    w/o GPU & 645.24 &	139.26 &	229.31 &	15.78 &	242.35	&11.51&	12.65\\
    w/ GPU & - & -& -& 1.35&	7.04&	5.57&	0.16\\
    \bottomrule[1pt]
    \end{tabular}
    }
    \end{table*}  
    
\subsubsection{Network architecture}
We use a  multi-scale network \cite{unet2015ronne} with shared weights to estimate depth maps from $\widehat{\mathcal{L}}_i$ $(i=1,2,3,4)$.
To be specific, we stack the SAIs in $\widehat{\mathcal{L}}_i$ along the feature channel, and feed them into the network.
The features extracted from the SAI stack are gradually down-sampled using max pooling layers with a  kernel of size  $2\times2$, and then gradually up-sampled to the original spatial size using transposed convolutional layers.
The feature numbers increase from $64$, $128$, $256$ to $512$ with the decrease of the spatial size.
In each scale, we use one layer of convolution and two residual blocks \cite{residual2016he} for feature extraction, and the features in the same scale are skip-connected to enhance the information flow.
Finally, in the up-sampling branch, we apply two convolutional layers to output the depth map and its corresponding reliability map at each scale.

The multi-scale structure improves the receptive field of the network, so that it can cover the correspondences on LFs with relatively larger disparity ranges. Moreover, the multi-scale network helps to propagate the depth estimation from rich-texture to weak-texture regions.

    \begin{figure*}[!t]
    \centering
    \includegraphics[width=\linewidth]{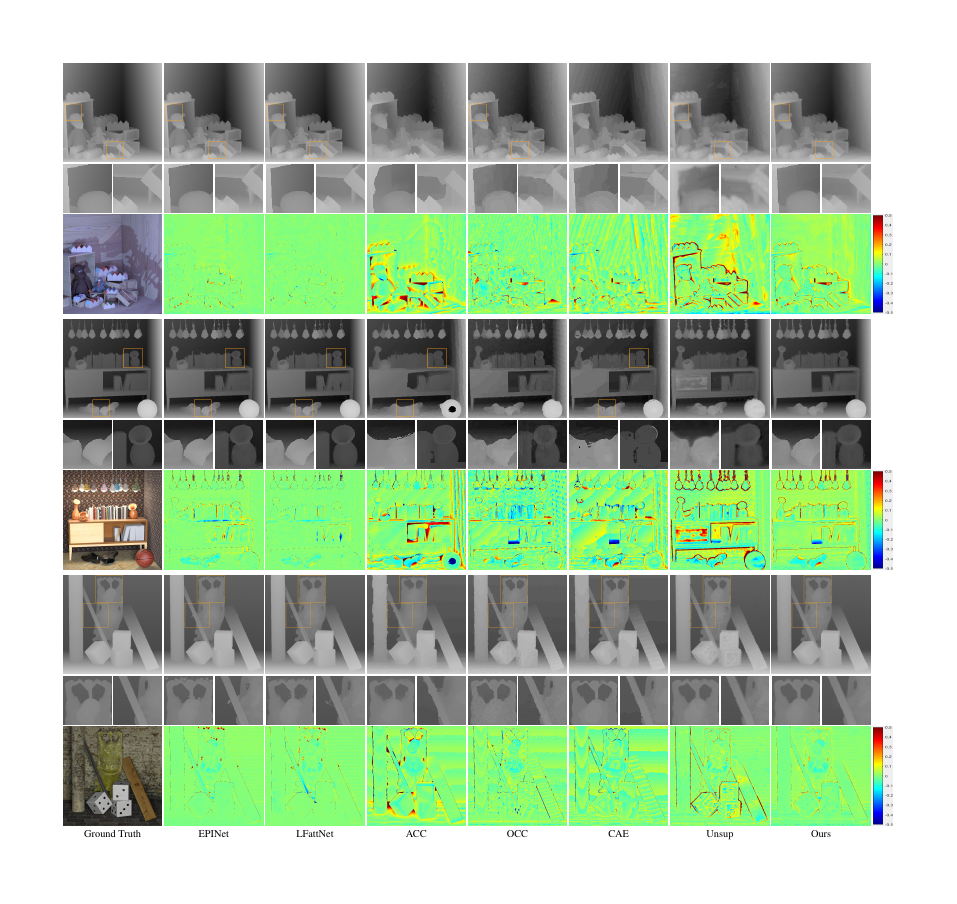}
    \caption{Visual comparison of the depth maps estimated by different methods on synthetic LF data from \cite{lfdataset2016hci} and \cite{lfdataset2013hci_old}. Selected regions have been zoomed in for better comparison. The error maps with respect to the ground-truth depth maps are also provided.}
    \label{fig_visual_syn}
    \end{figure*}
    
\subsubsection{Loss function}

To improve the smoothness of the estimated depth map, as well as encourage  the depth discontinuity on edges of the central view, we adopt an edge-aware smoothness loss 
\cite{tomasi1998bilateral,jonschkowski2020matters}, written as 
\begin{equation}
    \begin{aligned}
        \ell_{sm} = \frac{1}{2} \sum_{\mathbf{x}} \textsf{exp}\left(-\gamma\left|\frac{\partial I_{\mathbf{u}_0}}{\partial x}(\mathbf{x})\right| \right)\left|\frac{\partial \widetilde{D}}{\partial x}(\mathbf{x})\right| \\
        + \textsf{exp}\left(-\gamma\left|\frac{\partial I _{\mathbf{u}_0}}{\partial y}(\mathbf{x})\right| \right)\left|\frac{\partial \widetilde{D}}{\partial y}(\mathbf{x})\right|,
    \end{aligned}
\end{equation}
where the edge weight $\gamma$ is set to 150 empirically.

The final loss function is defined as $\ell = \ell_{c-rec} + \beta\ell_{sm}$, where $\beta$ controls the trade-off between the fidelity and  the smoothness. We empirically set $\beta$ to 0.1.

\subsubsection{Training details}
During training, we randomly cropped LF patches of spatial size $128\times128$ and angular size $7\times7$ from the training dataset.
We set the batch size to 4 and initialized the learning rate as $1e^{-4}$. 
We used Adam optimizer \cite{kingma2014adam} with $\beta_1 =0.9$ and $\beta_2 = 0.999$. 

\section{Experiments}
\label{sec:experiment}

To evaluate the performance of the proposed method, we compared with state-of-the-art algorithms, including three non-learning-based methods, i.e., ACC \cite{lfdepth2015accurate-jeon}, OCC \cite{lfdepth2015occlusion-wang}, and CAE \cite{lfdepth2017robust-williem}, two supervised learning-based methods, i.e., EPINet \cite{lfdepth2018epinet-shin} and LFattNet \cite{lfdepth2020attention-tsai},  and one unsupervised learning-based method, i.e., Unsup \cite{lfdepth2018unsupervised-peng} .

We performed comparisons on both synthetic and real-world LF data.
For the synthetic dataset, we  used 4 LF images from the HCI benchmark \cite{lfdataset2016hci} and 5 LFs from the HCI old benchmark \cite{lfdataset2013hci_old} for inference.
All learning-based methods were trained with the 4-D LF images from \cite{lfdataset2016hci} which are not included in the test set.
For the real-world LF data, we used 4-D LF images captured with a Lytro illum provided by Stanford Lytro LF Archive \cite{lfdataset2016stanford_lytro}, Kalantari \textit{et al.} \cite{lfdataset2016kalantari}, and EPFL LF dataset \cite{lfdataset2016epfl}.
As the real-world datasets have no ground-truth depth for supervised training, we only retrained the models of 2 unsupervised learning-based methods using 100 LF images from \cite{lfdataset2016kalantari}, while directly applying the model of supervised learning-based methods trained with synthetic data.
We also performed comparisons on the Stanford Gantry dataset \cite{lfdataset2016stanford_gantry}, which contains real-world LF images captured with a camera gantry.
For these datasets, we used the central $7\times7$ SAIs of the LF for depth estimation.

Besides, we also submitted our results to the HCI online benchmark\footnote{https://lightfield-analysis.uni-konstanz.de/benchmark/table} for evaluation, which is composed of eight synthetic LF images.  In addition to the aforementioned methods under comparison, we compared with two more state-of-the-art methods, i.e.,  one latest supervised learning-based method named AttMLFNet \cite{lfdepth2021attmlf-chen}  and one unsupervised learning-based method named Mono \cite{lfdepth2020monocular-zhou}\footnote{Note that as the source codes of AttMLFNet and Mono are not publicly available when submitting this manuscript, we did not compare our method with them on other datasets.}. Note that the results of all the compared methods in terms of this online benchmark are online available.

\subsection{Evaluation on Synthetic Data}
The available ground-truth depth of the synthetic data \cite{lfdataset2016hci,lfdataset2013hci_old} allows us to 
compare different methods quantitatively. Specifically, 
We computed  Mean Square error (MSE) and Bad Pixel Ratio (BPR) between the estimated depth maps and the ground-truth ones to measure the accuracy, i.e.,
\begin{equation}
\begin{aligned}
      MSE &= \frac{1}{N} \sum_{\mathbf{x}} (\widetilde{D}(\mathbf{x}) - D(\mathbf{x}))^2,\\
      BPR &=  \mathsf{Card}\left(\left\{\mathbf{x}: \left|\widetilde{D}(\mathbf{x})-D(\mathbf{x})\right|>t\right\}\right) / N,
\end{aligned}
\label{eq_metric}
\end{equation}
where $\widetilde{D}$ and $D$ are the estimated and ground-truth depth maps, respectively,
$N$ is the number of pixels in a depth map, and $\mathsf{Card}(A)$ denotes the element number of the  set $A$.
We set $t=0.07$ for evaluation.
Note that to avoid the padding issue in some methods, we cropped 20 pixels for the depth maps when computing the evaluation metrics.

Tables \ref{tab:quan_mse} and \ref{tab:quan_bpr} list the MSE and BPR values of different methods on HCI and HCI old datasets, and Tables \ref{tab:benchmark_mse} and \ref{tab:benchmark_bpr} 
on the HCI online benchmark, where we can observe that
\begin{itemize}
    \item As ground-truth depth maps are not utilized to train the model, the performance of unsupervised learning-based methods  is still obviously lower than 
    that of supervised methods on the synthetic dataset. Moreover, due to the lack of the consideration of 
    the occlusions, previous unsupervised learning-based method \cite{lfdepth2018unsupervised-peng} performs much worse than non-learning-based methods.
    
    \item Compared with 
    non-learning-based methods \cite{lfdepth2015accurate-jeon,lfdepth2015occlusion-wang,lfdepth2017robust-williem}, supervised learning-based methods \cite{lfdepth2018epinet-shin, lfdepth2020attention-tsai,lfdepth2021attmlf-chen} can achieve  superior performance on the HCI dataset, but the accuracy decreases seriously on the HCI old dataset. The reason may be that the image characteristics have great differences between these two datasets. Such a degradation also indicates that the domain difference between the training and testing data can limit the performance of the supervised learning-based methods.

    \item Our method produces results with the accuracy comparable to non-learning-based 
    methods \cite{lfdepth2015accurate-jeon, lfdepth2015occlusion-wang, lfdepth2017robust-williem}. Compared with supervised learning-based methods \cite{lfdepth2018epinet-shin,lfdepth2020attention-tsai,lfdepth2021attmlf-chen}, our method still has a performance gap on the HCI dataset, but can produce results with lower MSE values on the HCI old dataset.
    \item  Our method can achieve much higher accuracy than the unsupervised learning-based method  \cite{lfdepth2018unsupervised-peng}. Compared with the unsupervised learning-based method \cite{lfdepth2020monocular-zhou}, our method performs better in terms of MSE values and produces comparable BPR values on the \textit{Test} set, but worse on the \textit{Stratified} set. The possible reason is that the \textit{stratified} set contains LF images greatly different from real-world scenes to pose specific, isolated challenges, while our model is not explicitly designed to handle them.
\end{itemize}

     \begin{figure}[t]
     \centering
     \includegraphics[width=\linewidth]{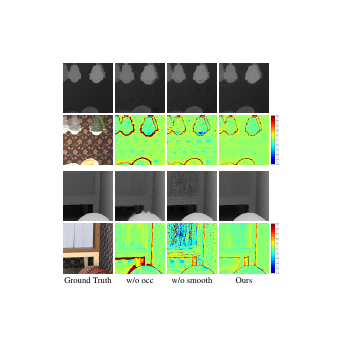}
     \caption{Visual lillustration of the effectiveness of the occlusion-aware strategy and the smoothness loss. 
     \textit{w/o occ} denotes the results produced without using the occlusion-aware strategy, and \textit{w/o smooth} means that the weight for the smoothness loss was set to 0.
     The depth maps and the corresponding error maps are provided. 
     }
     \label{fig_visual_ablation}
     \end{figure}  
     
    \begin{table}[t]
    \centering
    \caption{Ablation study for the proposed method. 
    \textit{w/o occ} denotes the results produced without using the occlusion-aware strategy, and \textit{w/o smooth} means that the weight for the smoothness loss was set to 0.
    The quantitative results (MSE $\times100$) of the depth maps are provided. 
    }
    \label{tab:ablation}
    \resizebox{0.8\linewidth}{!}{
    \begin{tabular}{c| c c c c }
    \toprule[1pt]
    ~ & Boxes & Cotton & Dino & Sideboard \\
    \midrule[0.5pt]
    w/o occ & 10.55 & 2.84 & 1.16 & 3.67 \\
    w/o smooth & 11.30 & 1.28 & 0.91 & 2.45 \\
    Ours & 7.45 & 0.80 & 0.63 & 1.79 \\
    \bottomrule[1pt]
    \end{tabular}
    }
    \end{table} 
    
Besides, Fig. \ref{fig_visual_syn} visualizes the depth maps predicted by different algorithms as well as their error maps compared with the ground-truth ones. 
It can be seen that our method can well handle most occlusion areas, and produce globally smooth depth maps, demonstrating the effectiveness of the proposed occlusion-aware strategy and the multi-scale network with the smoothness loss.

     \begin{figure*}[!t]
    \centering
    \includegraphics[width=\linewidth]{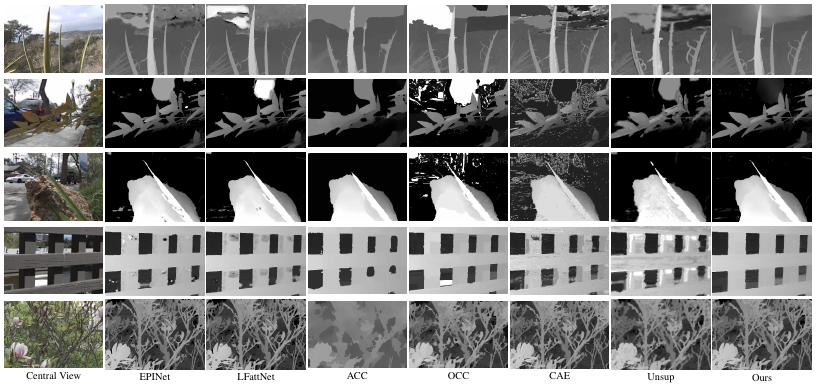}
     \caption{Visual comparison of the depth maps estimated by different methods on LF data from \cite{lfdataset2016stanford_lytro} and \cite{lfdataset2016kalantari}.}
     \label{fig_visual_lytro}
     \end{figure*}

     \begin{figure*}[!t]
     \centering
     \includegraphics[width=\linewidth]{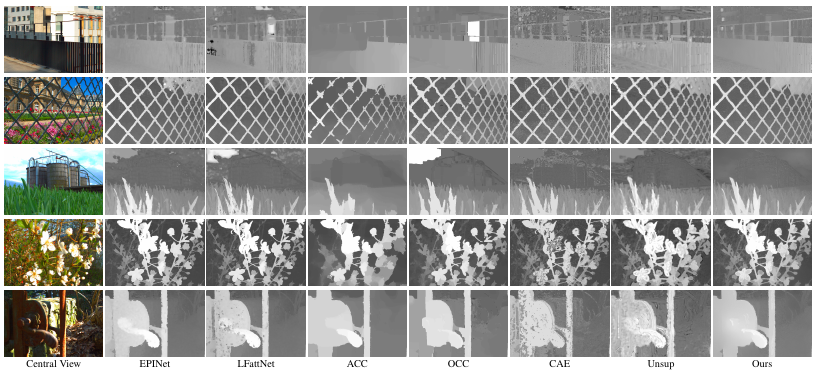}
     \caption{Visual comparison of the depth maps estimated by different methods on LF data from \cite{lfdataset2016epfl}. 
     }
     \label{fig_visual_epfl}
     \end{figure*}  

     \begin{figure*}[!t]
     \centering
     \includegraphics[width=\linewidth]{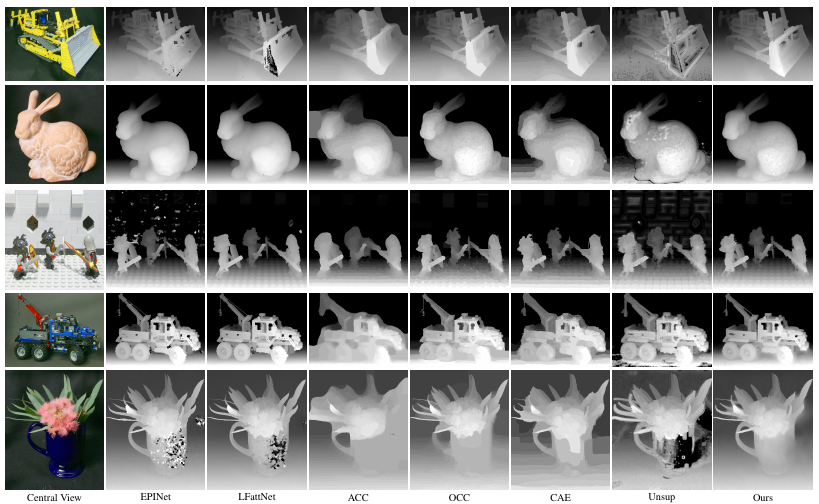}
     \caption{Visual comparison of the depth maps estimated by different methods on LF data from \cite{lfdataset2016stanford_gantry}}. 
     \label{fig_visual_gantry}
     \end{figure*}  
     
\subsection{Evaluation on Real-World Data}
We also compared the accuracy of estimated depth maps by different methods on real-world 4-D LF images \cite{lfdataset2016kalantari, lfdataset2016stanford_lytro} and \cite{lfdataset2016epfl}.
As the ground-truth depth maps are not available, 
we only visually compared the results. Figs. \ref{fig_visual_lytro}, \ref{fig_visual_epfl}, and \ref{fig_visual_gantry} present the estimated depth maps, 
where we can observe that although supervised learning-based methods \cite{lfdepth2018epinet-shin, lfdepth2020attention-tsai} can produce superior results on synthetic data, their performance is degraded on real-world LF images, especially on textureless areas.
The reason might be that the camera and environmental noises exist in the real-world images.
Traditional non-learning-based methods require many hyper-parameters for optimization, and thus, it is difficult for them to perform well on different scenes.
Additionally, although the unsupervised learning-based method \cite{lfdepth2018unsupervised-peng} was retrained on the real-world dataset, it produces depth maps with obvious inaccuracy on occlusion regions and textureless areas due to the intrinsic limitation of the model.
In comparisons, our method can produce satisfactory and better depth maps on various scenes.

\subsection{Comparison of Efficiency}

We compared the efficiency of different methods measured by the running time for inferring the depth map from a 4-D LF image, and Table \ref{tab:time} lists the results.
We provided the running time with and without the GPU acceleration.
All CPU-based methods were implemented  on a desktop with Intel CPU i7-8700 @ 3.70GHz, 32 GB RAM, and the GPU versions were accelerated by a NVIDIA Tesla V100.
From Table \ref{tab:time}, we can see that traditional non-learning methods suffer from the huge burden of computational costs, while learning-based methods, including EPINet \cite{lfdepth2018epinet-shin}, Unsup \cite{lfdepth2018unsupervised-peng}, and Ours, significantly save the running time even without the GPU acceleration. Particularly, our method is obviously faster than others under the acceleration of GPU.


    \begin{table*}[h]
    \centering
    \caption{Quantitative investigation of the effect of 
    the threshold $\lambda$ in occlusion detection on performance. The best results are highlighted in bold.}
    \label{tab:lambda}
    \begin{tabular}{l| c c c c | c c c c  }
    \toprule[1pt]
    $\lambda$ & Boxes & Cotton & Dino & Sideboard & Boxes & Cotton & Dino & Sideboard \\  
    \midrule[0.5pt]
    ~ & \multicolumn{4}{c|}{MSE $\times100$} & \multicolumn{4}{c}{BPR ($>0.07$)}  \\ 
    \midrule[0.5pt]
    0.1 & 7.53 & 0.83 & 0.73 & 2.14 & 32.15 & 9.91 & 15.23 & 26.23  \\ 
    0.2 & 7.46 & \textbf{0.80} & \textbf{0.62} & 1.89 & \textbf{26.08} & \textbf{8.33} & 8.28 & 15.26  \\ 
    0.3 & \textbf{7.45} & \textbf{0.80} & 0.63 & \textbf{1.79} & 26.24 & 8.46 & \textbf{8.25} & \textbf{14.20}  \\ 
    0.4 & 7.48 & 0.81 & 0.64 & \textbf{1.79} & 26.53 & 8.59 & 8.36 & 14.51  \\ 
    0.5 & 7.55 & 0.84 & 0.66 & 1.82 & 26.66 & 8.71 & 8.46 & 14.89  \\ 
    \bottomrule[1pt]
    \end{tabular}
    \end{table*}

    \begin{table*}[h]
    \centering
    \caption{Quantitative comparisons (MSE $\times 100$ and BPR ($>0.07$)) of using $D_{max}$ only and using both $D_{max}$ and $D_{avg}$. }
    \label{tab:max_only}
    \begin{tabular}{l| c c c c | c c c c  }
    \toprule[1pt]
    ~ & Boxes & Cotton & Dino & Sideboard & Boxes & Cotton & Dino & Sideboard \\  
    \midrule[0.5pt]
    ~ & \multicolumn{4}{c|}{MSE $\times 100$} & \multicolumn{4}{c}{BPR ($>0.07$)}  \\ 
    \midrule[0.5pt]
    $D_{max}$ only & 7.61 & 0.92 & 0.82 & 2.28 & 40.59 & 17.46 & 22.35 & 38.03  \\ 
    $D_{max}$ and $D_{avg}$ (Ours) & \textbf{7.45} & \textbf{0.80} & \textbf{0.63} & \textbf{1.79} & \textbf{26.24} & \textbf{8.46} & \textbf{8.25} &  \textbf{14.20} \\
    \bottomrule[1pt]
    \end{tabular}
    \end{table*}
    
\subsection{Ablation Study}

We also conducted ablation studies towards the proposed occlusion-aware strategy and the smoothness loss, and Table \ref{tab:ablation} and Fig. \ref{fig_visual_ablation} provide the quantitative and qualitative results, respectively. 

\subsubsection{The occlusion-aware strategy}
To demonstrate the advantage of the proposed occlusion-aware strategy, we developed a baseline named as \textit{w/o occ}, which directly predicts the depth map from the full LF image instead of the sub-LFs, and trained it by replacing the constrained photometric reconstruction loss in Eq. (\ref{eq_cons_recloss}) with the unconstrained one defined in Eq. (\ref{eq_unsuploss}).
From Table \ref{tab:ablation}, we can see that the accuracy of the predicted depth maps obviously decreases without using the occlusion-aware strategy.
Fig. \ref{fig_visual_ablation} shows that applying the occlusion-aware strategy significantly reduces the prediction error around occlusion boundaries.
 
\subsubsection{The smoothness loss}
To demonstrate the effectiveness of the smoothness loss, we trained a model named as \textit{w/o smooth} by setting the weight $\beta$ for the smoothness loss to 0. Experimental results presented in Table \ref{tab:ablation} show that the accuracy of the depth estimation seriously decrease without using the smoothness loss. From Fig. \ref{fig_visual_ablation}, we can observe that \textit{w/o smooth} has difficulties to estimate the depth in textureless areas, while Ours can produce accurate and consistent results by using the smoothness loss.

\subsubsection{The threshold $\lambda$ for occlusion detection}
To justify our selection of the value of the threshold $\lambda$, we quantitatively compared the predicted depth maps when setting different values of $\lambda$. 
As listed in Table \ref{tab:lambda}, we can see that setting $\lambda$ to 0.3 produces the depth maps with highest accuracy on most cases, while setting $\lambda$ to 0.1 or 0.5 causes obvious performance degradation.

\subsubsection{The necessity of using $D_{avg}$}
To quantitatively demonstrate the necessity of using $D_{avg}$, we compared the accuracy of predicted depth maps when 
using $D_{max}$ only and using both $D_{max}$ and $D_{avg}$. As shown in Table \ref{tab:max_only}, it can be seen that only using $D_{max}$ indeed reduces the accuracy of the estimated depth maps.

\section{Conclusion}
\label{sec:conclusion}

We have presented an unsupervised learning-based method for depth estimation from 4-D LFs. 
By introducing an occlusion-aware strategy incorporated with a constrained unsupervised loss and utilizing a multi-scale network with the smoothness loss, our method is capable of handling both the occlusion and textureless areas that are challenging for depth estimation, and produce satisfactory results on both synthetic and real-world LF images.
Experimental results demonstrate that our method can reduce the computational burden of traditional non-learning-based methods, alleviate the performance limitation of previous unsupervised methods, and overcome the problem of domain shift on supervised methods.



\ifCLASSOPTIONcaptionsoff
  \newpage
\fi



%



\bibliographystyle{IEEEtran}
\bibliography{IEEEabrv,./references}

\end{document}